\documentclass[letterpaper, 10 pt, conference]{ieeeconf}  

\IEEEoverridecommandlockouts                              

\usepackage[natbib=true,
            style=numeric-comp,
            sorting=none,
            backend=biber,
            ]{biblatex}
\usepackage{common}
\usepackage{this_doc}

\title{\LARGE \bf
Nested Sampling for Non-Gaussian Inference in SLAM Factor Graphs
}

\author{Qiangqiang Huang$^{1}$,
Alan Papalia$^{1, 2}$, and John J. Leonard$^{1}$
\thanks{Research supported by ONR grant N00014-18-1-2832,
ONR MURI grant N00014-19-1-2571, and the MIT-Portugal
program.}
\thanks{$^{1}$Computer Science and Artificial Intelligence Lab (CSAIL),
Massachusetts Institute of Technology, 77 Massachusetts Ave, Cambridge MA 02139, USA
{\tt\small \{hqq, apapalia, jleonard\}@mit.edu}
}%
\thanks{$^{2}$Department of Applied Ocean Physics and Engineering, Woods Hole
Oceanographic Institution, 86 Water St, Woods Hole, MA 02543, USA} %
}

\begin{document}
\maketitle
\thispagestyle{empty}
\pagestyle{empty}
\begin{abstract}
    We present nested sampling for factor graphs (NSFG), a novel nested sampling
    approach to approximate inference for posterior distributions expressed over
    factor-graphs. Performing such inference is a key step in simultaneous
    localization and mapping (SLAM). Although the Gaussian approximation often
    works well, in other more challenging SLAM situations,  the posterior
    distribution is non-Gaussian and cannot be explicitly represented with
    standard distributions. Our technique applies to settings where the
    posterior distribution is substantially non-Gaussian (e.g., multi-modal) and
    thus needs a more expressive representation. NSFG exploits nested sampling
    methods to directly sample the posterior to represent the distribution
    without parametric density models. While nested sampling methods are known
    for their powerful capability in  sampling multi-modal distributions, the
    application of the methods to SLAM factor graphs is not straightforward.
    NSFG leverages the structure of factor graphs to construct informative prior
    distributions which are efficiently sampled and provide notable
    computational benefits for nested sampling methods. We compare NSFG to state-of-the-art sampling
    approaches and Gaussian/non-Gaussian SLAM techniques in experiments. NSFG performs most robustly in describing non-Gaussian posteriors and computes solutions over an order of magnitude faster than other sampling approaches. We believe the primary value of NSFG is as a reference solution of posterior distributions, aiding offline accuracy evaluation of approximate distributions found by other SLAM algorithms.
\end{abstract}
\section{Introduction}

Simultaneous localization and mapping (SLAM), a fundamental problem in mobile
robotics, is commonly posed as inferring the posterior distribution of
robot and landmark states from relative measurements between the robots and
landmarks. These posterior distributions are often expressed with factor
graphs, which highlight the distribution's conditional independence structure \cite{bishop2006pattern}. Precise distribution estimation of the posterior
enables accurate uncertainty quantification for robust machine perception and lays a foundation for active perception
tasks that require decision-making under uncertainty. Due to nonlinearities in
real-world sensing models, the SLAM posterior is in general non-Gaussian. Exact
inference over those non-Gaussian posteriors is generally intractable so
practitioners resort to either deterministic or stochastic approximations.

\begin{figure}[!tbp]
    \centering
    \includegraphics[width=0.9\linewidth]{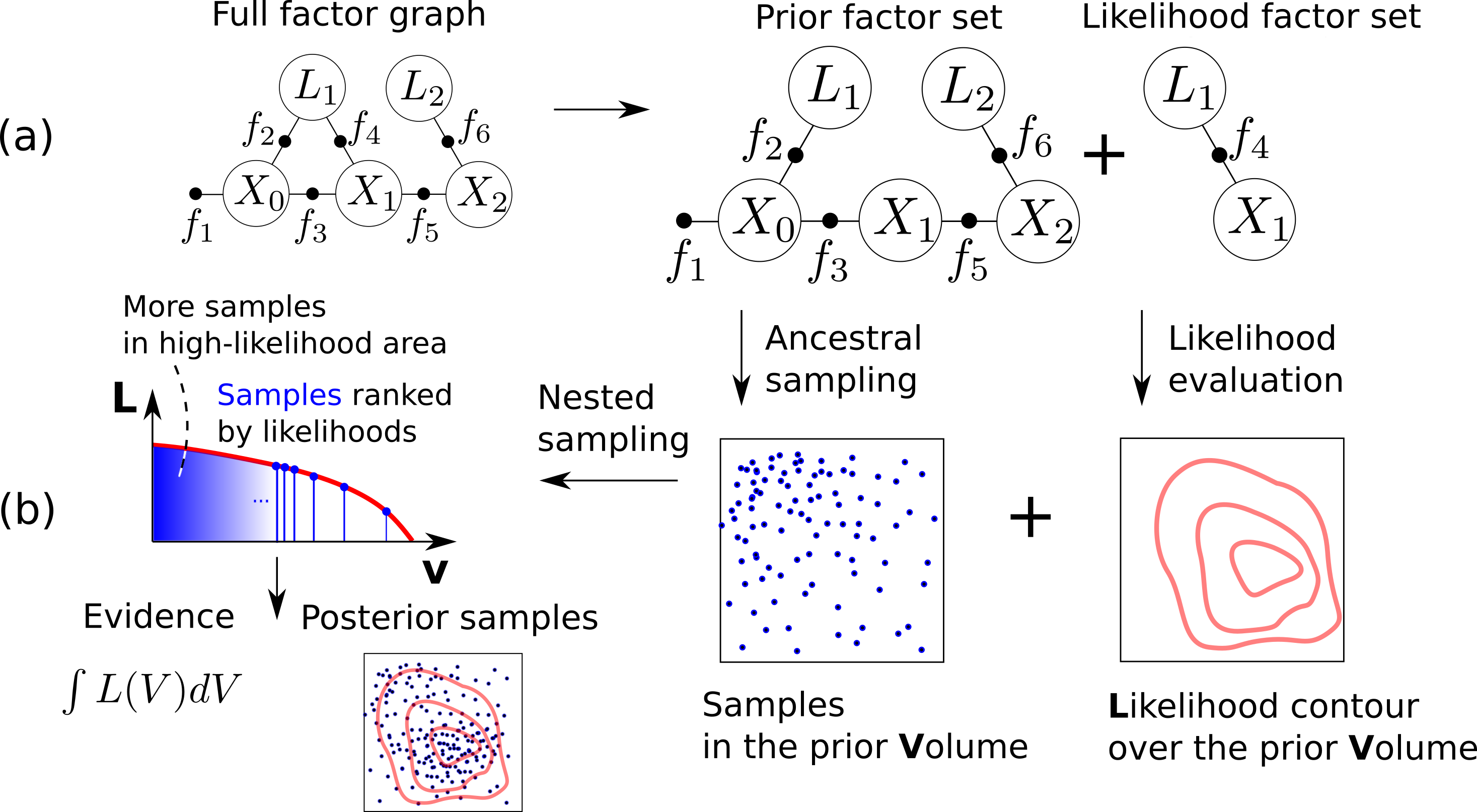}
    \caption{Our approach to performing nested sampling over factor graphs. (a) The factor graph is first decomposed into two parts: a prior factor set and a likelihood factor set.
    (b) From the acyclic graph in the prior factor set we apply ancestral sampling to generate samples of the prior distribution. The functions for sampling and likelihood evaluation are supplied to a nested sampler. Nested sampling returns estimated evidence and samples which approximate the posterior distribution.}
    \label{fig: factor graph}
    \vspace{-6mm}
\end{figure}

In cases where the posterior distribution is sharply peaked at a single point in
the state-space, it can be reasonable to approximate the distribution as Gaussian
through the Laplace approximation~\cite{bishop2006pattern}. However, the
Laplace approximation cannot represent the highly non-Gaussian posteriors which
arise in many SLAM problems due to circumstances such as range measurements,
multi-modal data association or loop closure, and non-Gaussian noise models. Thus many non-Gaussian SLAM algorithms have been proposed to pursue a tractable but more expressive approximation of the non-Gaussian posterior.

Stochastic approximation algorithms, which represent the posterior distribution
with samples, are generally recognized as more computationally expensive, but
more expressive, inference solutions~\cite{bishop2006pattern}. Our work falls
into this class of algorithms. The resulting samples of these algorithms enable qualitative analysis of the posterior as well as estimation
of statistical quantities of interest. Thus, stochastic approximation algorithms provide value in (1) offline accuracy evaluation of other
algorithms and (2) high-fidelity posterior estimation for problems which do not require real-time results onboard a robotic system.

Nested sampling is a recent stochastic approximation technique that is powerful for sampling multi-modal distributions \cite{skilling2006nested}. This work combines nested sampling with informative priors obtained from factor graphs. We term our algorithm nested
sampling for factor graphs (NSFG). To the
best of our knowledge, there are no existing general purpose algorithms for
reference solutions even on posteriors of small SLAM problems. This need for accuracy evaluation motivated the development of
NSFG, which pursues the \textit{bona fide} shape of the posterior and
thus aids accuracy evaluation of other SLAM inference algorithms.

NSFG is evaluated across four classes of simulated problems and a real-world dataset. Existing sampling approaches and state-of-the-art Gaussian/non-Gaussian SLAM algorithms are also compared. All implementations and experiments presented in this paper have been
made freely available\footnote{\RepoURL}.
\section{Related Work}

There are two key aspects to our work: the use of the factor graphs to improve
the underlying inference algorithm and the use of nested sampling to perform
approximate stochastic inference. We will first survey inference techniques in
SLAM with a focus on those leveraging factor graph structures. We will then
discuss general stochastic inference techniques which have not yet been applied
to SLAM.

\subsection{Factor graphs and inference in SLAM}

Factor graphs are a standard representation of SLAM problems which encode the
conditional independence structure of the posterior distribution
\cite{dellaert2017factor,rosen2021advances}. This structure is commonly
exploited to build efficient deterministic inference algorithms designed for
near-Gaussian posteriors
\cite{dellaert2006square,kaess2008isam,Kaess11icra,Kuemmerle11icra,Kaess12ijrr}.
From these solutions the posterior distribution can be approximated as a
Gaussian via the Laplace approximation \cite{kaess09ras,kaess2008isam}. Other
works extended the use of deterministic inference to
conditions such as false data associations \cite{sunderhauf12iros},
multi-hypothesis estimation~\cite{hsiao19icra}, and Gaussian
mixtures \cite{pfeifer2021advancing}. NSFG differs from these methods,
as it is sampling-based (i.e. stochastic inference) and does not assume any form
on the posterior distribution.


Beyond deterministic inference, the SLAM community has explored stochastic
inference techniques such as particle filtering
\cite{gonzalez09ras,blanco10ijrr,montemerlo2003fastslam} and Markov Chain Monte
Carlo (MCMC) \cite{shariff15aistats,Kaess05icra}. Many particle filtering
algorithms for SLAM (e.g., FastSLAM2.0~\cite{montemerlo2003fastslam}) track
landmarks by extended Kalman filters which cannot represent general
non-Gaussian/multi-modal distributions. Recent work leveraged factor graphs and
probabilistic modeling techniques such as non-parametric belief
propagation~\cite{Fourie16iros,fourie20iros} and normalizing
flows~\cite{huang21icra}. NSFG differs from these works in that it does not assume any parametric density models and builds upon a
stochastic inference technique (i.e. nested sampling) which is generally more
accurate for multi-modal distributions~\cite{speagle2020dynesty}. NSFG aims to improve inference fidelity at the
cost of computational complexity. This tradeoff suggests that NSFG could be used
in accuracy evaluation of approximate distributions found by SLAM techniques.

\subsection{General stochastic inference techniques}

Hamiltonian Monte Carlo (HMC) is an MCMC variant which guides sampling
exploration with the gradient of the density function
\cite{betancourt18conceptual}. The No-U-Turn Sampler (NUTS)
\cite{hoffman2014jmlr} extends HMC with automatic parameter tuning, improving
usability and performance. However, multi-modal distributions with distant modes
still pose general difficulties as state space exploration uses only local
density information. Specialized MCMC algorithms designed for multi-modal
distributions were recently
developed~\cite{tak2018repelling,pompe2020framework}.

Sequential Monte Carlo (SMC) is a sampling algorithm which combines importance
sampling, re-sampling, tempering, and MCMC. Particle filters in general are just
special instances of SMC \cite{doucet2009tutorial}. The use of tempering and MCMC alleviates sample impoverishment in standard particle filters (e.g., sampling importance resampling algorithms in~\cite[Alg. 4]{arulampalam2002tutorial}). SMC usually requires a
proposal distribution that possesses good coverage of the typical set of the
target density. It is difficult to design such a proposal distribution for a general
high dimensional target density.

Nested sampling was proposed by Skilling \cite{skilling2006nested} to compute
the evidence or marginal likelihood for Bayesian inference with a by-product of
posterior samples. It is mostly developed and used in the field of astronomy.
Nested sampling has two attractive features: (1) well-defined stopping criteria
related to the convergence of estimated evidence and (2) global exploration of
the state space. Nested sampling (NS) methods keep exploring and drawing new
samples from the state space until the estimate of evidence converges. NS has
demonstrated great
success with complex
posterior distributions that possess multiple modes~\cite{buchner2021nested}. The practical benefits of NS methods include: (i) no
predetermined number of samples and (ii) little to no tuning of proposal
distributions is required to get accurate results. The estimated evidence by
nested sampling is also helpful for model selection problems (e.g., multiple
factor graphs under different data associations). There are several open-source
nested sampling packages \cite{feroz2009multinest, handley2015polychord,
speagle2020dynesty}. In particular, \citet{speagle2020dynesty} developed
{dynesty}, an open-source dynamic nested sampling package which our
implementation of NSFG is built upon.
\section{Methods}
\label{sec: methods}

We define the SLAM inference problem that NSFG solves, discuss nested sampling
at a high level, and elaborate on how NSFG uses the conditional independence
structure of factor graphs to improve the abilities of nested sampling.

\subsection{Factor graphs and the problem formulation}
\label{sec:factor-graph}
We use the latent variable $\Theta$ to denote all robot poses ($X$) and landmark locations ($L$). Let $z$ be all measurements. {\it We aim to infer the posterior distribution of $\Theta$ given all measurements}, i.e.
$\PosteriorDensity$. The posterior distribution can be represented by a factor graph, as seen in \cref{fig: factor graph}a. A factor $f_i$ represents a likelihood function over one or more variables which are denoted by $\Theta_i$. We write the likelihood functions as $f_i(\Theta_i) =
    p(z_i|\Theta_i)$ in the case of measurements $z_i$ and as $f_i(\Theta_i) =
    p(\Theta_i)$ in the case of prior distributions. The joint posterior then relates to our factors as follows, with $m$ total
factors in the graph:
\begin{align}
    \PosteriorDensity \propto p(z|\Theta)p(\Theta) = \prod_{i=1}^m f_i(\Theta_i)
    \label{eq: posterior formation}
\end{align}




\subsection{Nested sampling}
\label{sec:nested-sampling}
Nested sampling \cite{skilling2006nested} was proposed to compute the evidence $p(z)$, with a by-product of posterior samples. The evidence is defined by an integral of the likelihood function $L(z|\Theta)$ and the prior distribution $\pi(\Theta)$ over the latent variable $\Theta$, as seen in
\begin{align} p(z) = \int _{\Theta} L(\Theta)\pi(\Theta)d\Theta, \label{eq:evidence}
\end{align}
where $L(\Theta)$ stands for the likelihood $L(z|\Theta)$. In nested sampling, samples of $\Theta$ are drawn from the
prior distribution. The prior distribution must cover all variables and is
ideally computationally efficient to sample. The likelihood function contains the
remaining factors of the posterior distribution.

Conceptually, nested sampling breaks up the sample space of the prior into \emph{nested} areas. Each nested area is enclosed by an iso-likelihood contour. We define the probability of a nested area as the prior volume $V \in [0, 1]$, while the likelihood on the contour is denoted by $L(V)$; thus, the small prior volume $dV$ denotes the probability of a small iso-likelihood shell in the sample space. With the notion of prior volumes, nested sampling transforms the integration \eqref{eq:evidence} to a one-dimensional integral of the likelihood over iso-likelihood small prior volumes, as follows~\cite{skilling2006nested, buchner2021nested}:
\begin{align} p(z) = \int
        _{0}^{1} L(V)dV. \label{eq: nested sampling foundation}
\end{align}

In practice, each of the samples from the prior represents a small prior volume $dV$, and the sum of all
small prior volumes is 1. Thus, the integration \eqref{eq:
        nested sampling foundation} can be implemented numerically by the
likelihood-weighted sum of those prior volumes to estimate the evidence. To
increase the precision of the numerical integration, nested sampling chooses to
draw more samples from the prior volume where the likelihood is higher. This can
be visualized using the $V$-$L$ plot in Fig.~\ref{fig: factor
        graph}b. Each bin under the $V$-$L$ curve corresponds to a sample. The
small prior volume and likelihood of a sample, respectively, determine the width
and height of the bin. Those bins are ordered from left to right following a
descending order of sample likelihoods (or bin heights). Thus, more samples
naturally lead to finer bins that approach the theoretical $V$-$L$ curve better.

Instead of drawing all samples of $\Theta$ only once from the prior distribution, nested
sampling draws samples across iterations. With each iteration the feasible space
of new samples gradually shrinks to high-likelihood areas in the sample space, leading to an
efficient and accurate estimate of the evidence. This gradual focusing is often
referred to as likelihood-restricted prior sampling (LRPS)
\cite{buchner2021nested}, and is the most important step in nested sampling. We
stress that LRPS is different from proposals in particle filters. While particle
filters attempt to draw samples of a variable once from a proposal, the LRPS
draws new samples across iterations until the estimated evidence converges; additionally, the weight of each sample depends on not only the likelihood but also the small prior volume.  As the contribution of our work is not in performing nested sampling, but in using
conditional independence structures from factor graphs to more efficiently
prepare a problem for nested sampling, we refer interested readers to
\cite{skilling2006nested,speagle2020dynesty,buchner2021nested} for further
details.


\subsection{Nested sampling for factor graphs}
\label{sec: nested sampling for fator graphs}
\subsubsection{Proposed approach}
While nested sampling is a powerful approach to sampling complex distributions,
naive application of nested sampling to SLAM does not take advantage of the
conditional independence structure of SLAM. We focus on how to use this
structure to enable nested sampling to be effectively applied to SLAM.
Specifically, we exploit the sparsity structure of SLAM factor graphs to
construct a prior $\pi(\Theta)$ and likelihood function $L(\Theta)$, which
enhances nested sampling.


Factors in SLAM factor graphs usually consist of a few unary factors as
priors, a number of binary factors modeling measurement likelihoods, and a
few factors connected to a robot pose and multiple landmarks for modeling
multi-model data association. As described in \cref{sec:nested-sampling}, the
prior model for nested sampling, $\pi(\Theta)$, must be a tractable distribution
from which samples of all latent variables can be efficiently drawn. Thus,
$\pi(\Theta)$ must incorporate factors more than the nominal priors to cover all
variables and, accordingly, these factors will be excluded from the likelihood
model, $L(\Theta)$. We will introduce our strategy for selecting factors that
compose $\pi(\Theta)$ and $L(\Theta)$.

Our strategy constructs a $\pi(\Theta)$ that enables ancestral
sampling~\cite{bishop2006pattern} for all variables, as ancestral sampling
admits very efficient distributional sampling. NSFG effectively builds
$\pi(\Theta)$ from a spanning tree, for trees naturally afford ancestral
sampling.

NSFG assumes that any variable in the SLAM factor graph is connected to at least
a binary factor (i.e., each variable is created along with a bivariate
factor), implying that binary factors already form a connected graph of all
variables. NSFG designates a node or variable connected to a prior factor as the
root of the spanning tree. Starting from samples drawn from the prior factor
for the root variable, one can use binary factors along the spanning tree to
generate samples of descendant variables up to the leaves of the tree. As seen
in Fig.~\ref{fig: factor graph}a, we designate the factors involved in this
ancestral sampling procedure as $\PF$ (prior factor set) and incorporate them in the
prior model $\pi(\Theta)$ while the remaining factors are referred to as $\LF$
(likelihood factor set) and make up the likelihood model, $L(\Theta)$. The resulting
factorization of the posterior distribution in (\ref{eq: posterior formation})
is \begin{align}
    \PosteriorDensity \propto \underbrace{ \prod_{f_j\in \LF} f_j(\Theta_j)
        }_{L(\Theta)} \underbrace{ \prod_{f_i\in \PF}f_i(\Theta_i)
        }_{\pi(\Theta)}\label{eq: posterior formation2}.
\end{align}
For example, for a typical SLAM factor graph with a single robot and $K$ unknown
landmarks, the $\PF$ set can be composed of the prior factor at the starting
pose of the robot, odometry factors, and $K$ binary factors that are connected
to different landmarks. Notions such as the $\PF$ set have commonly been used in
SLAM to initialize variables in MAP solvers and construct proposal distributions
in particle filters. This work introduces such approaches to nested sampling,
with a focus on sampling from joint posteriors in SLAM problems.



This partition of $\PF$ and $\LF$ sets improves nested sampling in two aspects:
(i) the prior model resembles the posterior distribution better than simple
proposals such as uniform distributions of all variables and (ii) the likelihood
model involves fewer factors, which reduces the cost of likelihood evaluation in
nested sampling. SLAM factor graphs are usually sparse, which implies that the
cardinality of $\PF$ set can be comparable to or even much greater than the
cardinality of $\LF$ set. Therefore, exploiting the sparsity structure of SLAM
factor graphs can effectively improve both the computational performance and
solution quality of nested sampling.

\subsubsection{Algorithms}
We implemented Algorithms~\ref{algo: structured factors} and \ref{algo:
nested-sampling-based SLAM} for obtaining the $\PF$ and $\LF$ sets and drawing posterior
samples via NSFG. The factors in the $\PF$ set are constructed from a
spanning tree, as seen in Algorithm~\ref{algo: structured factors}, and are thus
ordered for performing ancestral sampling. Note that selection of the prior
factor for designating the root of the tree in line~\ref{line:pick-prior} of
Algorithm~\ref{algo: structured factors} depends on user-defined heuristics
(e.g., choosing a more informative prior). Using the $\PF$ and $\LF$ sets,
Algorithm~\ref{algo: nested-sampling-based SLAM} defines the prior model
$\pi(\Theta)$ and likelihood model $L(\Theta)$ that enable nested sampling.
The likelihood model is simply evaluating the sum of the log-likelihoods of the
factors in the $\LF$ set (line~\ref{line:likelihood} in Algorithm~\ref{algo:
nested-sampling-based SLAM}). The use of the $\PF$ set for realizing
$\pi(\Theta)$ is less straightforward since it involves density transformation
(line~\ref{line:cube-sample} in Algorithm~\ref{algo: nested-sampling-based
SLAM}).

Nested sampling methods usually require transformation functions that map
from the uniform distribution over unit hypercube to the prior
distribution $\pi(\Theta)$, rather than explicit expressions of prior distributions 
~\cite{feroz2009multinest,handley2015polychord,speagle2020dynesty,buchner2021nested}. We refer to these transformations as hypercube transforms. Hypercube transforms
are necessary because the unit hypercube is first sampled to enable
global exploration of the state space. The transform applied to these
samples casts the global exploration into the domain of the variable $\Theta$, improving global coverage of the state space. Hypercube transforms can be implemented by quantile functions of noise models, which are available in typical SLAM problems (e.g., Gaussian noise models). With these hypercube transforms and observation models, we obtain samples for our first variable in the $\PF$ set and
then apply ancestral sampling to propagate the samples along the tree encoded in the $\PF$ set (line~\ref{line:prior-model} in Algorithm~\ref{algo: nested-sampling-based SLAM}).

\setlength\textfloatsep{10pt}
\setlength\floatsep{5pt}
\setlength\intextsep{10pt}
\begin{algorithm}[!t]
  \fontsize{9pt}{9pt}\selectfont
  \DontPrintSemicolon 
  \KwIn{Univariate factor set $\mathcal{P}$, binary factor set $\mathcal{B}$, and all other factors $\mathcal{L}$}
  \KwOut{Prior factor queue $\PF$ and likelihood factor queue $\LF$}

  Initialize empty FIFO queues of $\PF$ and $\LF$

  Construct a spanning tree $\mathcal{T}$ from the graph formed by binary factors $\mathcal{B}$

  Push a prior factor $f$ from $\mathcal{P}$ to $\PF$ and designate its variable as the root of $\mathcal{T}$ \label{line:pick-prior}

  Traverse $\mathcal{T}$ from the root to leaves and push binary factors to $\PF$ once they have been visited

  Push all factors that are not in $\PF$ to $\LF$

  \Return{$\PF$ and $\LF$}\;
  \caption{{Obtain $\PF$ and $\LF$ sets (spanning tree)}}
  \label{algo: structured factors}
\end{algorithm}


\begin{algorithm}[!t]
  \fontsize{9pt}{9pt}\selectfont
  \DontPrintSemicolon 
  \KwIn{Prior factor queue $\PF$ and likelihood factor queue $\LF$}
  \KwOut{Samples of the joint posterior distribution}
  \SetKwProg{Fn}{Function}{:}{}
  \Fn{LLK(latent variable $\Theta$) \label{line:likelihood}} 
  {
    $l\gets0$ \tcp*{Initialize log likelihood}
    \For{$f$ in $\LF$}{
      $l=l+f.\mathsf{loglikelihood}(\Theta)$\;
    }
    \KwRet{$l$}\;
  }
  \Fn{PriorTrans($\text{hypercube sample}$ $\bm{u}$)}
  {Initialize a dictionary $\mathcal{P}$ for containing prior samples\;
    \For(\tcp*[h]{Ancestral sampling in each iteration}){$f$ in $\PF$}{\label{line:prior-model}
      $v\gets$ \text{Variable in $f$ but not in }$\mathcal{P}$\;
      $\mathcal{P}[v] \gets f.\mathsf{hypercube\_transform}(\bm{u},\mathcal{P},v)$ \label{line:cube-sample}\;
    }
    \KwRet{$\mathcal{P}$}
  }
  $\mathcal{S}\gets NestedSampling(PriorTrans,LLK)$ \label{line:nested-sampling}\;
  \Return{$\mathcal{S}$}\tcp*{Return posterior samples}
  \caption{NSFG}
  \label{algo: nested-sampling-based SLAM}
\end{algorithm}

\section{Implementation}
\subsection{Observation and noise models}
\label{sec:noise-models}

We introduce observation and noise models that will be used in our
experiments. A noisy pose observation is defined by $\NoisyPose=\TruePose
\exp(\PerturbationHat)$ where pose $\TruePose \in \SE(d)$ is a latent variable,
$\wedge$ turns $\Perturbation$ into a member of the Lie algebra $\se (d)$, and
$\Perturbation \sim \Gaussian(\zeros , \Sigma)$ is the perturbation vector
subject to a Gaussian distribution. A noisy range measurement is modeled by
$\NoisyDist = \lVert \mathbf{t}_i - \mathbf{t}_j \rVert_2 + \Gaussian(0,
\sigma^2) \in \R$ where $\mathbf{t}_i$ is the translation component of variable
$X_i$ or $L_i$.
Beyond binary factors with known data associations, we use sum-mixture factors to model ambiguous data association as follows
\begin{align}
    f_i(\Theta_i)=p(z_i|\Theta_i)=\sum_{j=1}^{|\mathcal{D}|}p(z_i|\Theta_i,d_j)p(d_j) \label{eqn:sum-mixture-factor}
\end{align}
where each component is a binary factor with certain data association
$d_j\in{\mathcal{D}}$. In our experiments we assume the data association prior
$p(d_j)$ is a uniform distribution given no prior knowledge.

\subsection{Other solvers for comparison}
We use an open source package called dynesty, developed by \citet{speagle2020dynesty}, for performing
nested sampling in line~\ref{line:nested-sampling} of Algorithm~\ref{algo: nested-sampling-based SLAM}. A vanilla sampler based on nested sampling was also implemented to justify the
advantage of exploiting factor graph structure. In the vanilla sampler, all SLAM
factors are incorporated into the likelihood model for nested sampling while
predetermined uniform distributions of all variables are supplied as the prior
distribution. This vanilla sampler is denoted as
NS(UnifPr) in~\cref{sec:results} for comparison.

Two other state-of-the-art stochastic inference methods, NUTS and SMC, a
state-of-the-art Gaussian SLAM solver, GTSAM~\cite{dellaert2012factor}, and a
non-Gaussian SLAM solver, NF-iSAM~\cite{huang21icra}, are tested in our
experiments as well. We supply our SLAM factors to the NUTS and SMC
implementations in PyMC3~\cite{salvatier16probabilistic}, GTSAM, and NF-iSAM to
solve our SLAM problems. We used the default built-in initialization functions
in PyMC3 for NUTS and provided a predetermined uniform distribution that covers
the space of interest to SMC. The C++ library of GTSAM was used while all other
techniques were implemented in Python. All computation was run on an AMD Ryzen
ThreadRipper 3970X processor with 32 cores.
\section{Results}
\label{sec:results}
NSFG is evaluated across four simulated and one real-world dataset to observe
different aspects of its performance and capability. The datasets are:
pose-graph SLAM (\cref{sec:experiments-pose-graph}), range-only SLAM
(\cref{sec:experiments-single-robot-ro}), sensor network
localization (\cref{sec:experiments-snl}), ambiguous data association
(\cref{sec:experiments-single-robot-ambiguous-ro}), and the Plaza1 dataset (Sec.
\ref{sec:plaza1}). We emphasize that NSFG pursues high-fidelity samples of the posterior at the cost of computational complexity. While examples in this work
are small-scale, they possess abundant non-Gaussian features in posteriors,
warranting their potential as canonical examples for comparing algorithms in
this work and validating efficient non-Gaussian inference techniques in the
future.

Qualitative evaluation is performed by plotting the samples drawn by each
inference algorithm. Points of different colors indicate different variables. We
choose to compare the empirical mean, rather than the MAP point, of the samples with the ground truth to compute the
RMSE. This choice was made for two reasons: (i) the MAP point among samples can be very random in posteriors with equally weighted modes, and (ii) the MAP point does not reflect secondary modes when gauging distributional errors.

\subsection{Pose graphs}
\label{sec:experiments-pose-graph}
\begin{figure}[!t]
    \centering
    \includegraphics[width=0.99\linewidth]{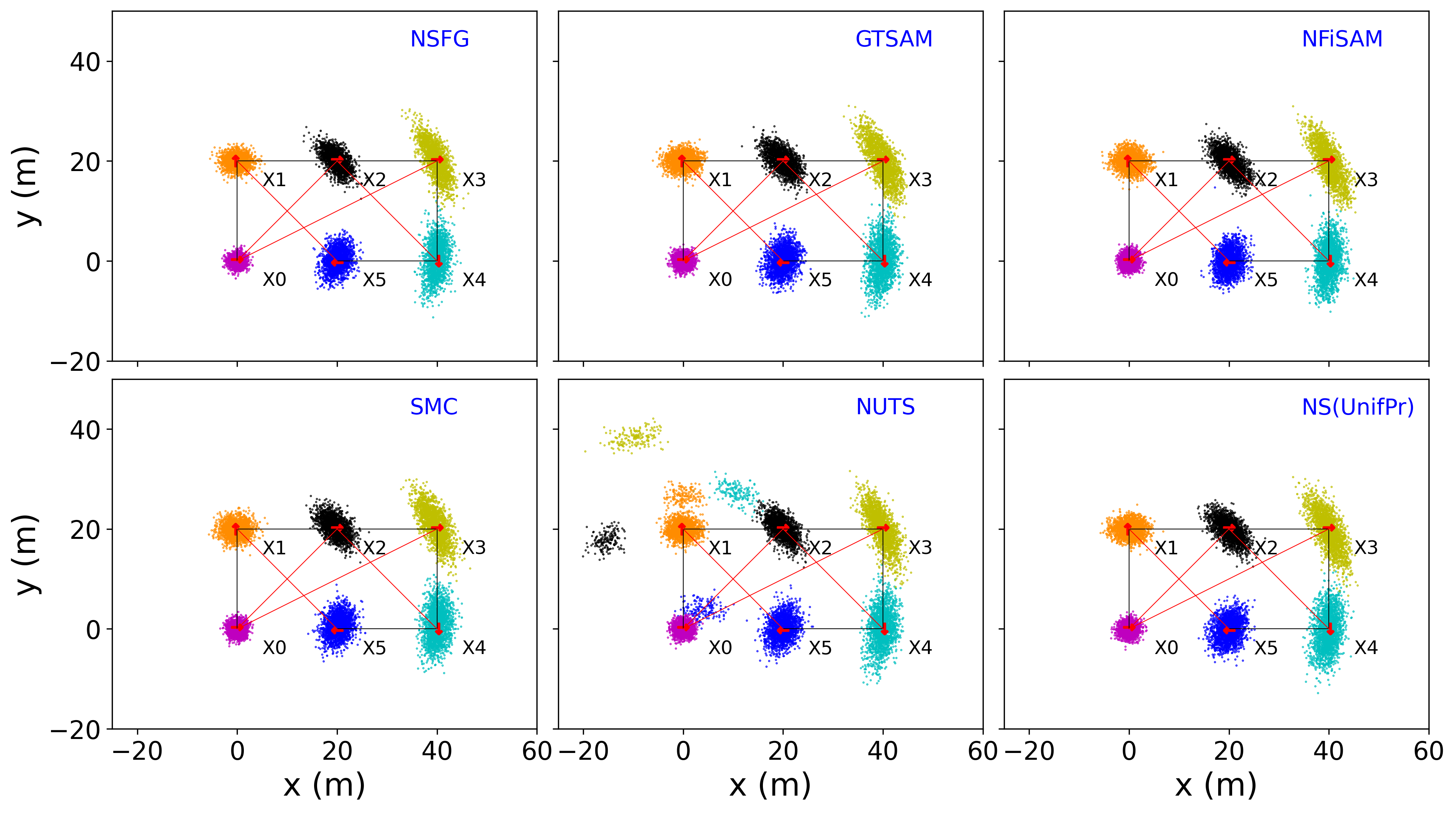}
    \caption{Samples that represent estimated posterior distributions for a
            pose-graph with 6 poses (X0-X5) and 4 loop-closures. The robot starts in the lower left and travels counter-clockwise. The groundtruth poses
            are marked by arrows. The black lines indicate odometry measurements between
            successive poses. The red lines indicate loop closures. Samples are marked
            by colored dots, with different colors indicating different pose variables.}
    \label{fig:PG_baseline_scatter}
\end{figure}

To test the simplest scenario in which all
methods would be expected to work, we first evaluate 2D pose graphs. In \cref{fig:PG_baseline_scatter} we show the results on one such problem.
Samples of the GTSAM solution are drawn from the Laplace approximation provided
by GTSAM. Estimated distributions by different methods were qualitatively
similar with the exception of some spurious modes found in the NUTS solution.

\begin{figure}[!h]
    \centering
    \includegraphics[width=0.8\linewidth]{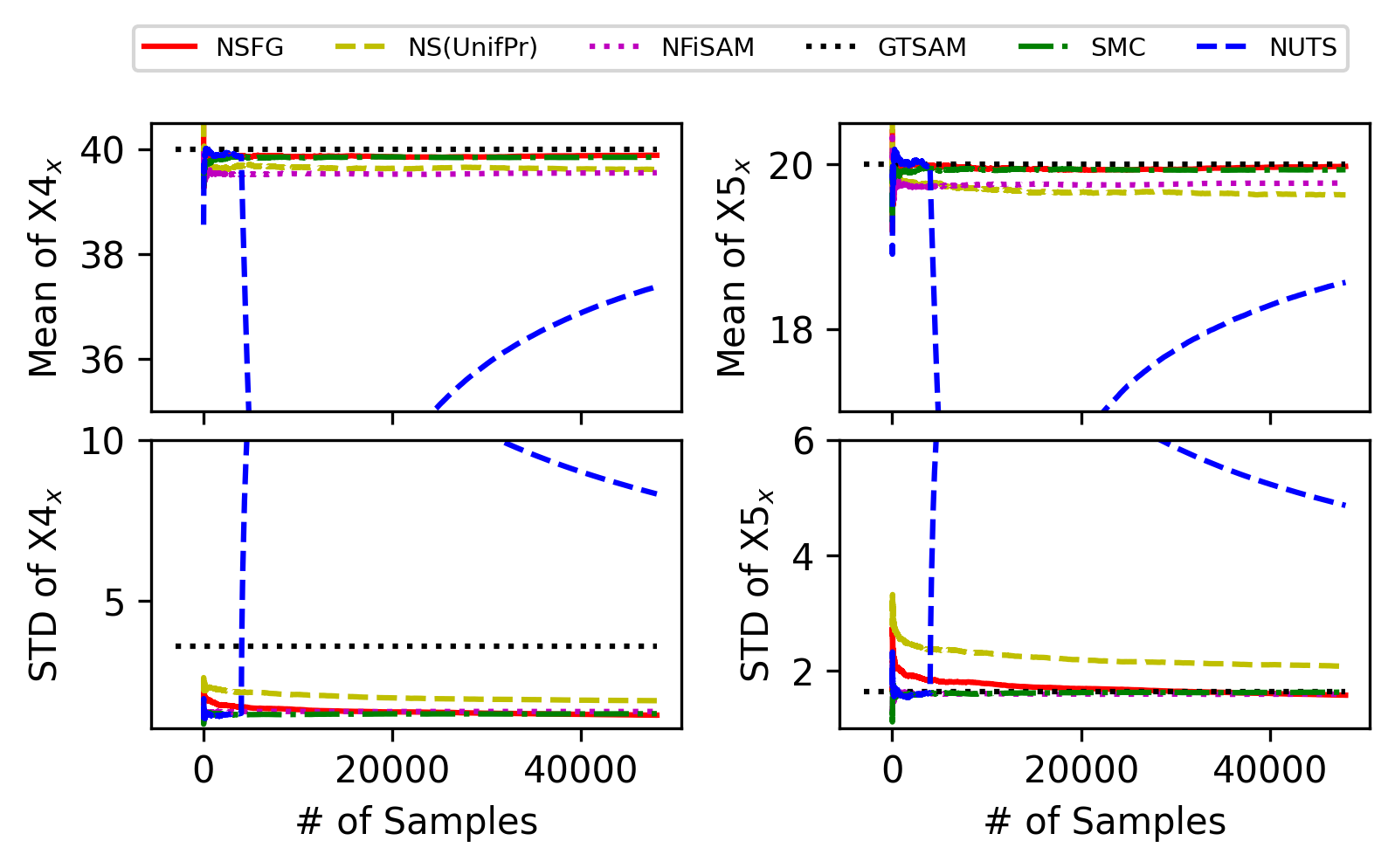}
    \caption{Empirical mean and standard deviation over samples of $x$ coordinates of selected variables ($X4$ and $X5$) in the pose graph example in Fig.~\ref{fig:PG_baseline_scatter}. Note that samples generated in the burn-in and tuning stages of NUTS are discarded.}
    \label{fig:convergence plot}
\end{figure}

\cref{fig:convergence plot} indicates that estimates made by the different samplers
indeed converge as more samples are drawn. As a result of spurious modes shown in~\cref{fig:PG_baseline_scatter}, the estimates by NUTS converge to different values from those by the other solvers. It is worth noting that estimated
standard deviations of the $x$ coordinate of $X4$ by NSFG, SMC, NS(UnifPr), and NFiSAM converge to roughly the same value, but visibly differ from the GTSAM estimate.  This
difference is reasonable since the standard deviation estimated by GTSAM is a
local Gaussian approximation at the MAP point of this non-Gaussian posterior.

\begin{figure}[!t]
    \centering
    \includegraphics[width=0.99\linewidth]{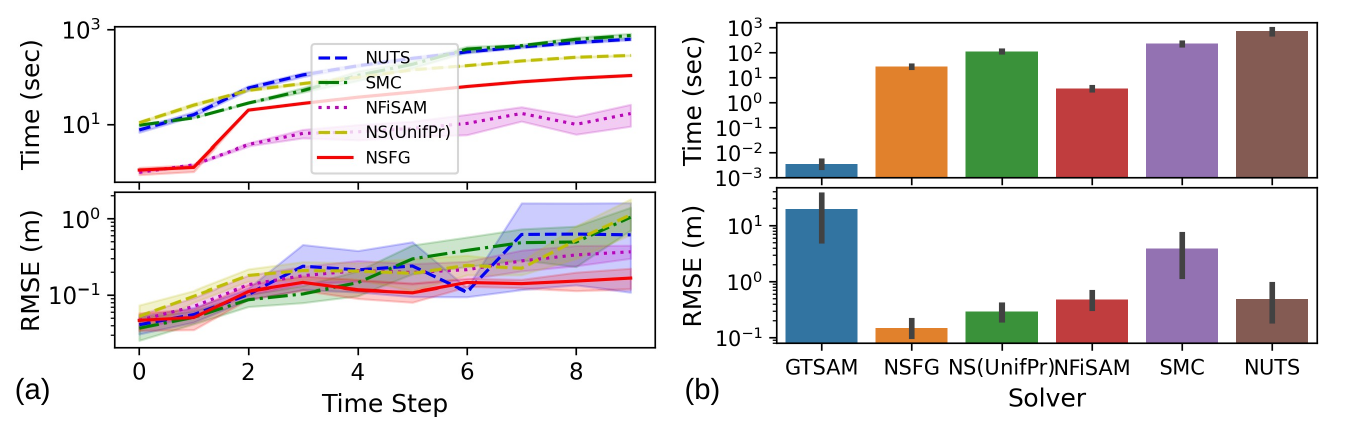}
    \caption{Performance of different approaches in two experiments: (a) 10 randomly generated pose graphs with 30 dimensions each (10 poses), and (b) 10 randomly generated range-only SLAM problems with 14 dimensions (4 poses and 1 landmark position). Shaded areas in plot (a) and error bars in plot (b) indicate the $95\%$ confidence interval.
        }
    \label{fig:all_baseline_performance}
\end{figure}

For quantitative evaluation, \cref{fig:all_baseline_performance}a shows the root
mean squared error (RMSE) and runtime of solutions for 10 randomly generated
pose graphs with 10 poses. The runtime and accuracy of GTSAM are not plotted in
\cref{fig:all_baseline_performance}a as GTSAM outperformed the RMSE and runtime
of other approaches by several orders of magnitude on these pose graphs. Since
the posteriors in these pose graphs are expected to be unimodal, a large RMSE in
the figure is intended to indicate samples from spurious modes. Compared with
the vanilla sampler relying on uniform prior distributions, NS(UnifPr), both
accuracy and runtime performances are improved in NSFG. Notably, NSFG appears
faster by a factor of 3-4. These results suggest the benefits of supplying an
informative prior model to nested sampling. The narrow error bands for NSFG suggest that NSFG is more robust than other samplers.


\subsection{Range-only SLAM}
\label{sec:experiments-single-robot-ro}

\begin{figure*}[!t]
    \centering
    \includegraphics[width=0.9\linewidth]{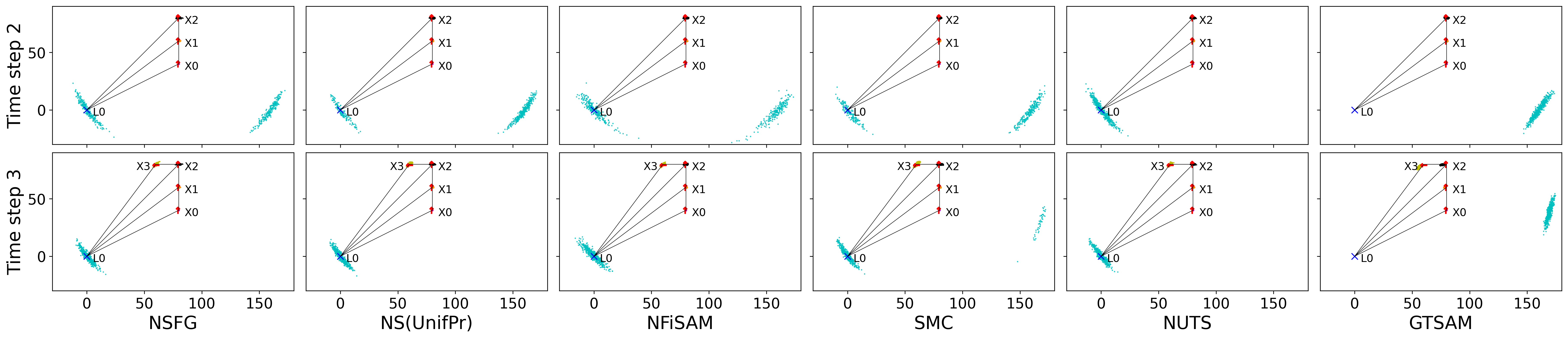}
    \caption{Solutions generated by each of the inference methods for a
            range-only SLAM problem. The red
            arrows indicate the groundtruth robot pose and the blue points
            indicate samples of the estimated landmark state. The red and black lines denote range and odometry measurements, respectively. For the GTSAM solution, the initial value of the landmark $L0$ is randomly picked on a circle that centers around the pose $X0$.}
    \label{fig:RO_baseline_scatter}
\end{figure*}

\begin{figure}[!htbp]
    \centering
    \includegraphics[width=0.85\linewidth]{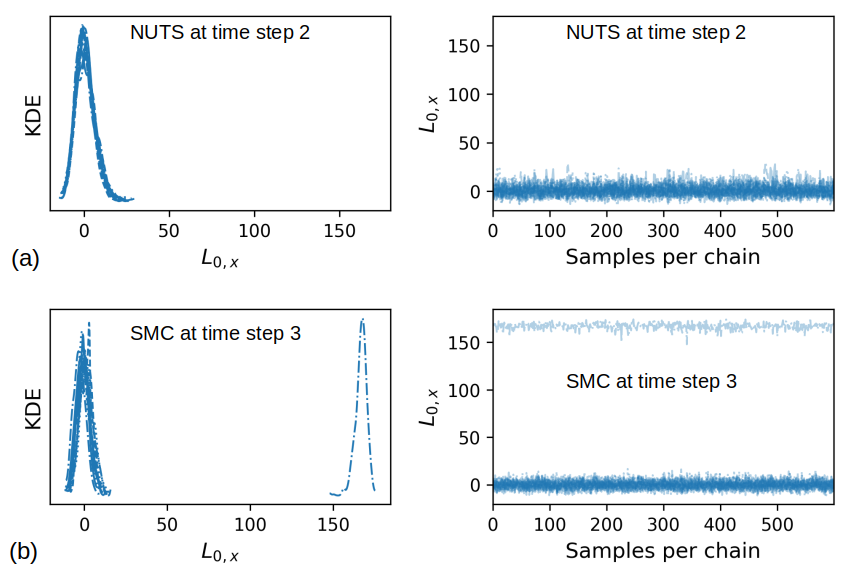}
    \caption{Traces of the x-coordinate values of the landmark explored by the
            MCMC chains in \cref{fig:RO_baseline_scatter} and kernel density estimation
            performed on the final distributions obtained from these MCMC chains. (a)
            NUTS solution for the baseline range-only problem at time step 2 and (b) SMC
            solution for the baseline range-only problem at time step 3.}
    \label{fig:RO_traceplot}
\end{figure}

We evaluate range-only SLAM problems to explore performance on a
simple, well-understood problem with non-Gaussian, multi-modal posterior
distributions. We show a single result from the range-only SLAM
problems in \cref{fig:RO_baseline_scatter}. In the first three time steps (X0,
X1, X2) the robot moves along a line, as such the posterior distribution of the
landmark position consists of two distinct modes mirrored across the line driven
by the robot.  At the final time step (X3) the robot breaks away from the line,
disambiguating the landmark position and causing the posterior distribution to
converge to a single mode around the true landmark position. Qualitative
evaluation of the solutions shows that NSFG best matches the expected posterior distribution for all time steps. This strongly
suggests that NSFG can best approximate the posterior distribution of range-only
SLAM problems.




In \cref{fig:all_baseline_performance}b we quantitatively compare all solvers on 10 other randomly generated range-only experiments. The
statistics presented were computed at the final time step of each experiment
after the distribution becomes unimodal (similar to time step 3 in
\cref{fig:RO_baseline_scatter}). NSFG enjoys the lowest RMSE across different approaches. Note that in both pose-graph and range-only SLAM experiments, NSFG presents advantages over the other sampling techniques for accuracy and runtime (faster by over an order of magnitude in~\cref{fig:all_baseline_performance}).

To explain the errors in SMC and NUTS, in \cref{fig:RO_traceplot} we display
diagnostics explaining the issues observed in \cref{fig:RO_baseline_scatter}. As
observed in the two plots on the right of \cref{fig:RO_traceplot}, the MCMC
chains are effectively stuck in local optima of the posterior. These local optima prevent the MCMC chains from mixing
between the two modes of the distribution, leading to an incorrect estimate of weights over different modes. In the case of the NUTS solution at
time step 2 (\cref{fig:RO_baseline_scatter}), all of the MCMC chains are stuck
around a single mode, and are thus prevented from exploring and sampling
the equally probable mirrored solution. In the case of the SMC solution at time
step 3 (\cref{fig:RO_baseline_scatter}), since there is a local optimum around the spurious landmark
position $L_{0,x} = 160$ which some MCMC chains cannot escape, samples from these chains incorrectly stress and overestimate the mode around the spurious landmark position.

\subsection{Sensor network localization}
\label{sec:experiments-snl}
\begin{figure}
    \centering
    \includegraphics[width=0.99\linewidth]{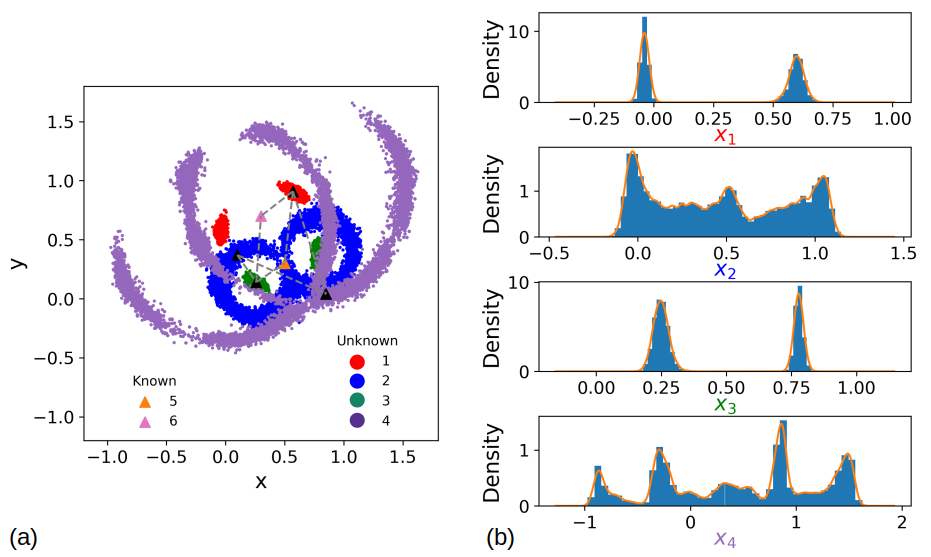}
    \caption{Samples and corresponding histograms and kernel density estimation drawn by NSFG for the sensor network localization example in~\cite{tak2018repelling}. The four black triangular markers designate the groundtruth locations of four sensors whose positions are initially unknown.  The two colored triangles designate the locations of two sensors at known locations.  The dashed lines indicate range measurements from the various sensors to one another.}
    \label{fig:SNL example}
\end{figure}

To evaluate NSFG on a well-known non-Gaussian inference problem, we tested
against the sensor network localization problem of~\cite{tak2018repelling}. The
scenario and solution by NSFG are shown in \cref{fig:SNL example}. In brief, the
inference goal is to estimate the posterior distributions of four unknown sensor
locations, $\mathbf{t}_1,\ldots,\mathbf{t}_4$, on the $x$-$y$ plane provided two
sensors with known locations, $\mathbf{t}_5$ and $\mathbf{t}_6$, a few range
measurements, $\{y_{ij}\}$, and a likelihood model. The
measurement likelihood between sensors $i$ and $j$ is modeled as
\begin{align*}
     & f_{ij}(\mathbf{t}_i,\mathbf{t}_j|y_{ij},w_{ij})= \\
     &
    \begin{cases}
        \exp(-\frac{\lVert \mathbf{t}_i-\mathbf{t}_j \rVert_2^2}{2 \times 0.3^2})\exp(-\frac{(y_{ij}-\lVert \mathbf{t}_i-\mathbf{t}_j \rVert_2)^2}{2 \times 0.02^2}), & \text{if}\ w_{ij}=1 \\
        1-\exp(-\frac{\lVert \mathbf{t}_i-\mathbf{t}_j \rVert_2^2}{2 \times 0.3^2}),                                                                                  & \text{otherwise.}
    \end{cases}
\end{align*}
where $w_{ij}$ equals 1 if there is a distance measurement between sensors $i$ and $j$ otherwise it is zero. Thus the posterior for this example is
\begin{align*}
    p(\mathbf{t}_1,\ldots,\mathbf{t}_4|\mathbf{y},\mathbf{w})\propto \prod _{\substack{i=2,\ldots,6 \\ j=1,\ldots,4 \\ i>j}}f_{ij}(\mathbf{t}_i,\mathbf{t}_j|y_{ij},w_{ij}).
\end{align*}

The scatter plots, histograms, and kernel density estimation in \cref{fig:SNL
        example}a and b highly resemble those in~\cite{tak2018repelling}, further
justifying the NSFG's ability to represent highly non-Gaussian posteriors.

\subsection{Range-only SLAM with ambiguous data association}
\label{sec:experiments-single-robot-ambiguous-ro}

\begin{figure}
    \centering
    \includegraphics[width=0.8\linewidth]{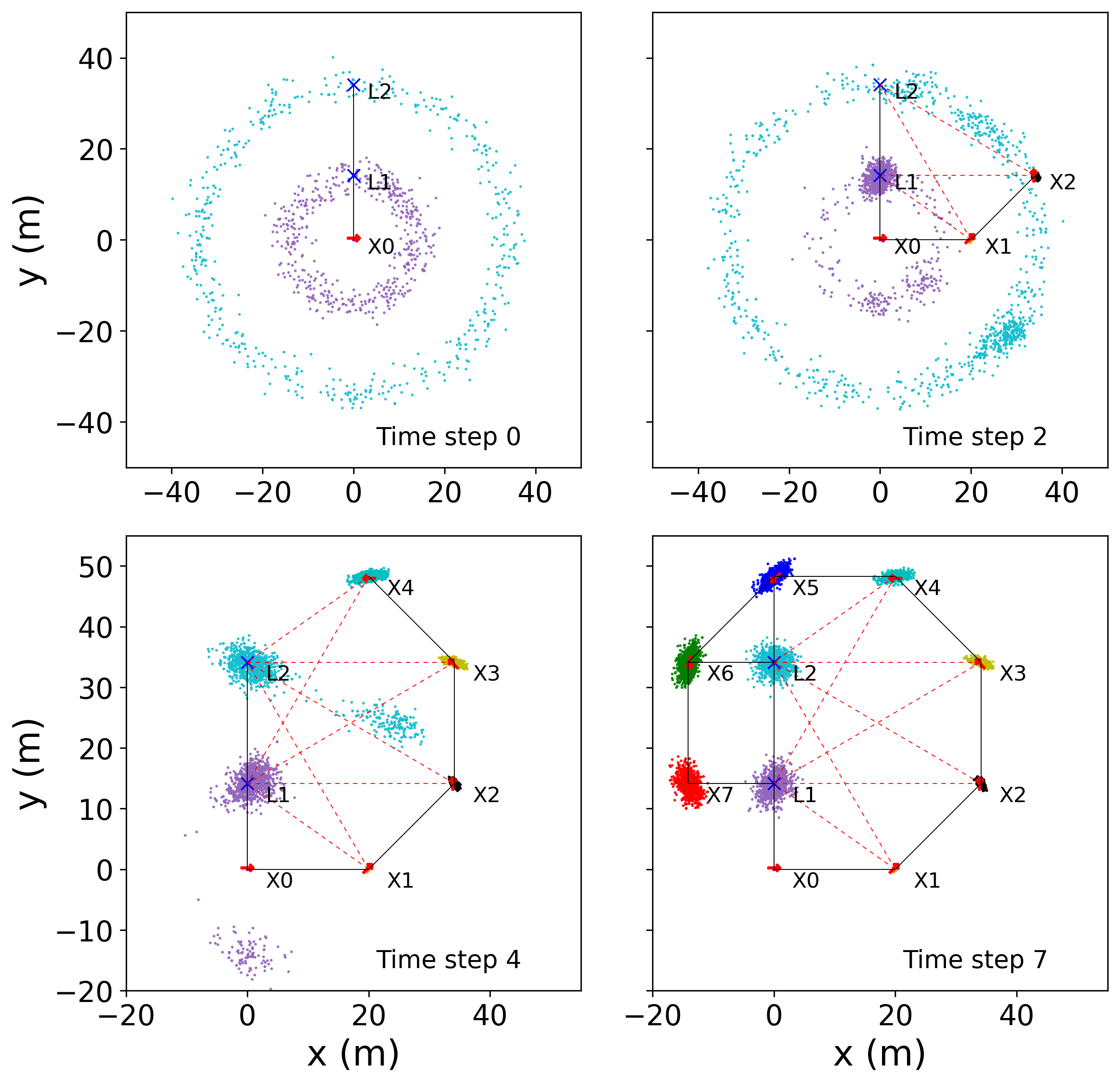}
    \caption{Scatter plots for posteriors of a range-only SLAM problem with data
            association ambiguity using NSFG. The dashed red lines from the same robot pose (X) to multiple landmarks (L) denote a range measurement that can be associated with all these landmarks. Black lines denote measurements with
            known association.}
    \label{fig:all_da_scatter}
\end{figure}

\begin{figure}[!htbp]
    \centering
    \includegraphics[width=0.75\linewidth]{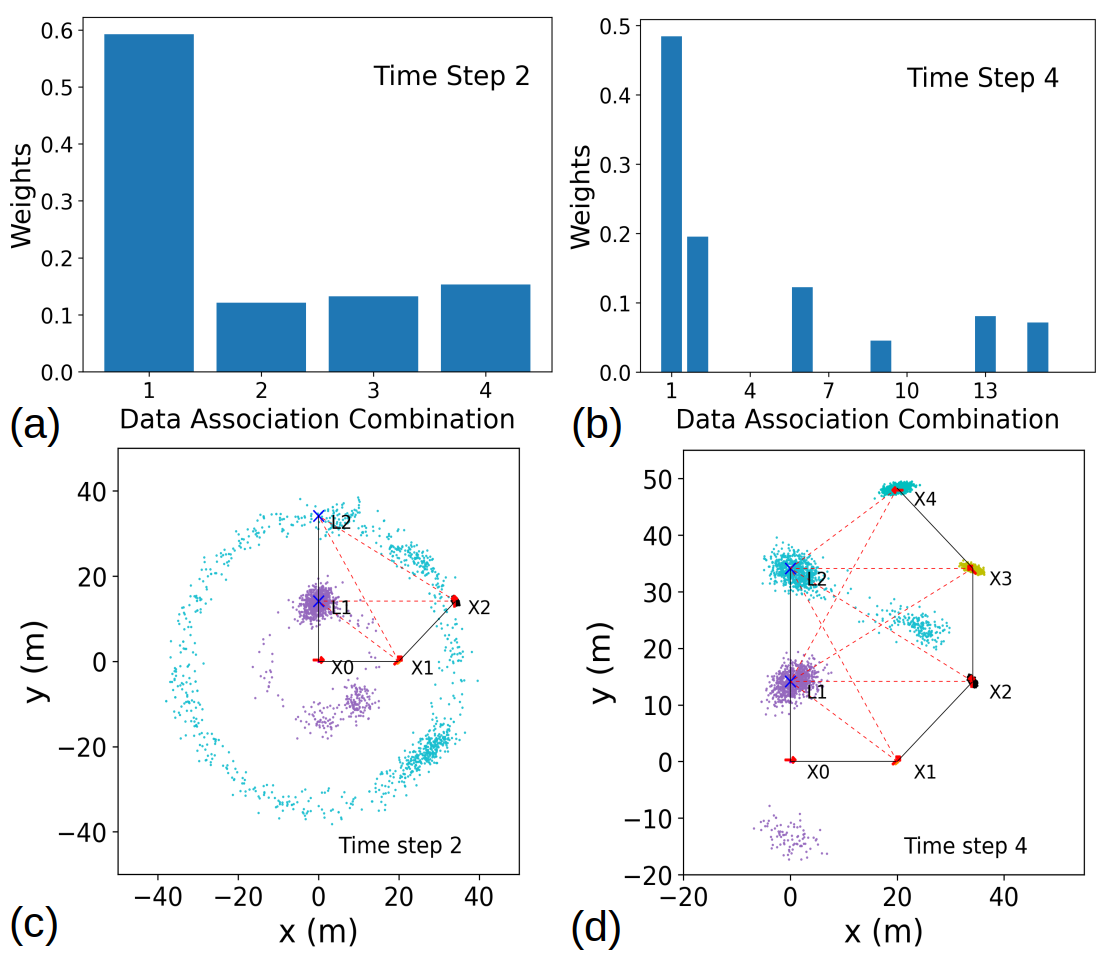}
    \caption{Weights and posteriors under different data associations. (a) and (b) are estimated weights of data association hypotheses at time step 2 and 4 respectively. (c) and (d) are posterior samples formed by combining samples from individual components in \cref{eq: data association formula}.}
    \label{fig:DA_weights}
\end{figure}

In \cref{fig:all_da_scatter,fig:DA_weights}, we evaluate the performance of
NSFG on a range-only SLAM problem with data association ambiguity. At time step 0, range measurements to two beacons are acquired with the identify information of beacons; from pose $X1$ to $X4$, however, the landmark association is
ambiguous (i.e. the robot is unsure of which landmark the distance measurement
goes to). This class of problems was chosen as the posterior distribution is
highly complex and demonstrates that NSFG can solve mixed continuous-discrete
inference problems. In ground truth, $L1$ is observed by $X1$ and $X2$ while $X3$ and
$X4$ spot $L2$.

As seen in \cref{fig:all_da_scatter}, NSFG displays the posterior distribution that arises from this situation and is
capable of disambiguating the landmark locations by time step 7.

Alternatively, the posterior with data-association ambiguity can be
represented with the following mixture model:
\begin{align}
    p(\Theta|z) & = \sum_{i=1}^{|\mathcal{D}|} w_i p(\Theta|z,d_i) \label{eq: data association formula}, \\
    w_i         & =\frac{p(\mathnormal{z}|d_i)p(d_i)}{\sum_{i=1}^{|\mathcal{D}|} p(z|d_i)p(d_i)},
\end{align}
where $\mathcal{D}$ denotes the set all data association combinations.
Each mixture component stems from one of the data association hypotheses, i.e.,
$p(\Theta|z,d_i)$.

Fixing the data association for a given combination, $d_i$, results in a new
posterior, $p(\Theta|z,d_i)$. For the new posterior, NSFG can draw samples and
estimate the evidence, $p(z|d_i)$. As there is no prior on data associations,
$p(d_i)$ is assumed to be
$\frac{1}{|\mathcal{D}|}$. Thus, the weights of components in (\ref{eq: data
        association formula}), $w_i$, can be computed if we apply NSFG to solve factor
graphs resulted from all combinations of data association
(\cref{fig:DA_weights}, top). A new ensemble of samples representing the joint
posterior can be formed by performing re-sampling over the samples and weights
for different data associations, as seen in the bottom of \cref{fig:DA_weights}.
These scatter plots resemble their counterparts in \cref{fig:all_da_scatter}
well. Effectively, this demonstrates that NSFG is self-consistent and can be
used in multiple ways to reliably approximate complex posterior distributions.

\subsection{Real-world dataset}
\label{sec:plaza1}
We apply NSFG to solve early time steps in the Plaza1
dataset~\cite{djugash14ijfr}. The dataset provides a sequence of timestamped
odometry and range measurements collected by a mobile robot in a planar
environment. The ranges were measured between the robot and four landmarks using
ultra-wideband sensors for which a noise model can be found
in~\cite{huang2021online}. We use measurements in the early stage of the
sequence to create a range-only SLAM factor graph. The factor graph involves 29
robot poses and 4 landmark locations, thus leading to a 95-dimensional posterior
distribution at the final time step. The posterior distribution contains
one prior factor, twenty-eight odometry factors, and thirty-three range factors.
Additionally, we generate three more synthetic problems from this dataset which
associate $20\%$, $40\%$, and $60\%$ of range measurements with all landmarks.
Such a range measurement with ambiguous data association (ADA) is modeled as a
sum-mixture \eqref{eqn:sum-mixture-factor}. These mixture factors are
expected to incur a higher function evaluation cost than binary factors (a
mixture factor entails evaluating 4 binary factors, given 4 landmarks here). We
use these synthetic problems to investigate the impact of complex likelihood
factor sets on the performance of NSFG.

\begin{figure}[!htbp]
    \centering
    \includegraphics[width=0.99\linewidth]{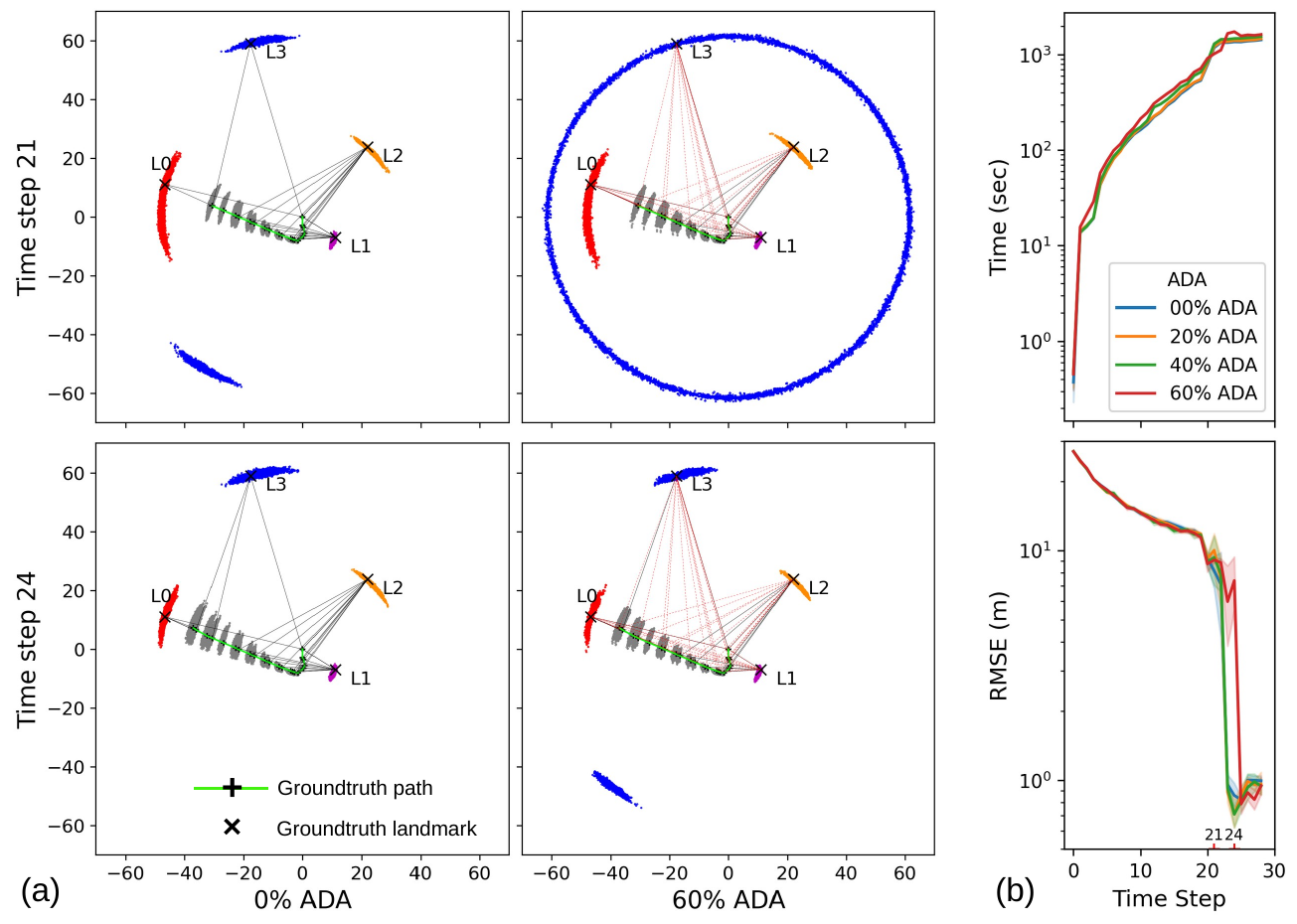}
    \caption{Plaza1 dataset with different fractions of ADA: (a) posterior samples and (b) runtime and error.
            In plot (a), black lines indicate range measurements with known data
            association. Red lines from a robot pose indicate ambiguous
            range measurements. In (b),
            the shaded area indicates the $95\%$ confidence interval estimated
            by six runs with different random seeds.}
    \label{fig:plaza1-scatter}
\end{figure}

Fig.~\ref{fig:plaza1-scatter}a shows samples drawn by NSFG at time steps 21 and
24. Given data association ($0\%$ ADA), we estimate that the posterior at time
step 21 is dominated by two modes, which differ in the location of landmark
$L3$; at time step 24, the posterior becomes unimodal. In contrast, the belief of landmark $L3$ becomes
much more uncertain in the $60\%$ ADA case. The RMSE in
Fig.~\ref{fig:plaza1-scatter}b supports this claim of distributional
uncertainty. The transition from multimodal to unimodal posteriors sharply
decreases the RMSE in all cases. However, such sharp reduction occurs to the
$60\%$ ADA case later than other cases. In the end of the sequence, the RMSE of
$60\%$ ADA converges to a low value, indicating the recovery of the groundtruth
mode. As seen in Fig.~\ref{fig:plaza1-scatter}b, more ADA factors incur greater
computational cost, but these factors do not alter the scaling behavior of
runtime with respect to dimensions. Note that later time steps entail more poses
and greater dimensionality. In addition, we solve each of the factor graphs six
times using different random seeds. The narrow error bands in
Fig.~\ref{fig:plaza1-scatter}b indicate that random seeds have little effect on
NSFG.

We note that NSFG confirms that the joint posterior of landmarks and the robot path
becomes peaked at a single point within the early time steps solved here. For
the subsequent time steps in the dataset, one can confidently apply the MAP estimation and (unimodal)
Gaussian approximation to represent the joint posterior.

\section{Conclusion and Future Work}
We introduced nested sampling methods to directly draw samples from the posterior distributions encountered in non Gaussian SLAM problems, pursuing the \textit{bona fide} shape of the posterior. Leveraging the
sparsity structure of SLAM factor graphs, our proposed approach, NSFG, provides nested sampling with informative prior distributions which can be efficiently sampled, leading to computational benefits for nested sampling methods.

We have demonstrated the advantage of NSFG over other sampling techniques and state-of-the-art Gaussian/non-Gaussian SLAM algorithms. NSFG presents superior robustness in inferring posteriors than all other approaches and operates over an order of magnitude faster than other
sampling techniques. Additionally, we showed that the estimate of evidence, as a unique benefit of nested sampling, can be used to compute the posterior belief of ambiguous data associations, indicating the potential of NSFG for solving mixed continuous-discrete inference problems. Lastly, the performance of NSFG was demonstrated to be consistent under various conditions such as dimensionality, fractions of ambiguous data associations, and random seeding.


We believe that NSFG can be a promising tool for providing reference solutions for the posteriors in non-Gaussian SLAM problems. These solutions can aid accuracy evaluation of
approximate inference algorithms and promote deeper understanding of uncertainty propagation on cyclic non-Gaussian SLAM factor graphs. Future work includes developing goodness-of-fit criteria of posterior samples for SLAM.



\balance
\printbibliography
\end{document}